\documentclass[conference]{IEEEtran}
\def\IEEEtitletopspace{18pt}
\IEEEoverridecommandlockouts
\usepackage{tikz}
\usepackage{pgfplots}
\pgfplotsset{compat=1.18} 
\usetikzlibrary{patterns,positioning}
\usepackage{multirow}
\usepackage{caption}
\usepackage{cite}
\usepackage{amsmath,amssymb,amsfonts}
\usepackage{algorithmic}
\usepackage{algorithm}
\usepackage{graphicx}
\usepackage{textcomp}
\usepackage{xcolor}
\def\BibTeX{{\rm B\kern-.05em{\sc i\kern-.025em b}\kern-.08em
    T\kern-.1667em\lower.7ex\hbox{E}\kern-.125emX}}

\usepackage{cuted}
\usepackage{subcaption}
\usepackage{array}
\usepackage{comment}
\usepackage{balance}

\newcolumntype{C}[1]{>{\centering\arraybackslash}m{#1}}
\renewcommand\IEEEkeywordsname{Keywords}
    
\begin{document}

\title{Scale-based Approach for Active\\Wildfire Segmentation on Satellite Imagery
\thanks{This research is sponsored by national funds through Fundação para a Ciência e a Tecnologia (FCT), under projects UID/00285/2025 (DOI: 10.54499/UID/00285/2025) and LA/P/0112/2020 (DOI: 10.54499/LA/P/0112/2020), and co-financed by the European Regional Development Fund through Innovation and Digital Transition Thematic Programme (COMPETE 2030) of Portugal 2030, Project FireSys (COMPETE2030-FEDER-02230700).}
}

\author{
\IEEEauthorblockN{
Matheus F. Kovaleski\IEEEauthorrefmark{1},
Cristiano Premebida\IEEEauthorrefmark{1},
João Ruivo Paulo\IEEEauthorrefmark{1}
}

\IEEEauthorblockA{\IEEEauthorrefmark{1}
Institute of Systems and Robotics, Department of Electrical and Computer Engineering, University of Coimbra, Portugal\\
Email: \{matheus.kovaleski, jpaulo, cpremebida\}@isr.uc.pt
}

}

\maketitle

\begin{abstract}
Active wildfire mapping from satellite imagery is challenging due to the sparse and highly imbalanced nature of fire pixels, especially in early-stage or low-density fire observations. This work investigates the use of multispectral Landsat-8 imagery for active-fire segmentation under multi-scale wildfire size conditions. We propose a data-driven protocol to characterize fire-region size distributions through connected-component analysis and an interquartile range criterion, enabling the evaluation of model robustness across different local fire-region densities. Three segmentation architectures, U-Net, DeepLabV3+, and SegFormer, are evaluated under different SWIR-based spectral configurations. Results show that U-Net achieves the strongest robustness across the evaluated conditions, SegFormer provides competitive performance, and DeepLabV3+ tends to produce conservative predictions with reduced recall. Across architectures, SWIR2 consistently achieves the strongest or near-best results, highlighting its importance for active-fire segmentation in Landsat-8 imagery. These findings suggest that both spectral band selection and architectural design are critical for robust satellite-based active wildfire mapping trained on low active fire-pixel density images.
\end{abstract}


\section{Introduction}
\label{sec.Introduction}

Wildfires are among the most significant natural disturbances affecting terrestrial ecosystems, with increasing frequency and intensity over recent decades due to climate change \cite{ellis2021global, senande2022global}. Large wildfire events produce severe environmental, economic, and social impacts, including biodiversity loss, carbon emissions, and risks to human life \cite{jones2022global}. In addition to detection and monitoring, recent research has also explored computational approaches for wildfire spread prediction and simulation under complex environmental conditions \cite{pereira2022wildfire, pereira2024metaheuristic}.

Satellite remote sensing has become an essential tool for large-scale wildfire monitoring, enabling near-real-time observation of extensive and remote areas using multispectral sensors such as Landsat-8, Sentinel-2, MODIS, and VIIRS \cite{pereira2021activefire}. In recent years, deep learning approaches have significantly improved wildfire segmentation performance, particularly through convolutional neural networks (CNNs) such as U-Net and DeepLabV3+ \cite{ronneberger2015unet, chen2018encoder} and more recent transformer-based architectures \cite{xie2021segformer}.

Despite these advances, wildfire segmentation remains challenging due to severe class imbalance, spectral ambiguity, and large variability in fire morphology and spatial extent \cite{tran2025land8fire}. In particular, wildfire events evolve from small ignition regions to large and spatially complex fire fronts, introducing significant changes in spatial patterns and contextual structure.

Most existing studies evaluate wildfire segmentation models using conventional random train-test splits \cite{tran2025land8fire, rashkovetsky2021wildfire, gao2025dafdm}. Although effective for measuring average performance, such protocols may not properly assess model robustness under distribution shifts caused by fire-scale variability.

In this work, we investigate wildfire segmentation under a scale-based distribution shift scenario, where models trained on small fire regions are evaluated on substantially larger active fire regions. To this end, we implement a data-driven train/test splitting strategy based on the interquartile range (IQR) of connected fire-region sizes extracted from the Land8Fire dataset. Furthermore, we analyze the influence of different SWIR-based spectral configurations on segmentation performance. We compare three segmentation architectures: U-Net, DeepLabV3+, and SegFormer.

The main contributions of this work are summarized as follows:

\begin{itemize}
    \item A scale-based dataset splitting strategy for evaluating wildfire segmentation robustness under fire-size distribution shifts;
    
    \item An analysis of the impact of SWIR spectral configurations on wildfire segmentation performance;
    
    \item A comparative evaluation of CNN-based and transformer-based segmentation architectures under shifted fire-scale distributions.
\end{itemize}

\section{Related Work}
\label{sec:related_work}

Wildfire detection using remote sensing has been widely studied, benefiting from the increasing availability of multispectral satellite imagery. Early approaches relied on threshold-based methods and spectral indices derived from thermal and shortwave infrared (SWIR) bands, exploiting the strong radiative response of active fires in these wavelengths \cite{pereira2021activefire, giglio2018modis}. While effective in controlled scenarios, these methods are often sensitive to atmospheric effects, noise, and spectral confusion with non-fire elements.

With the emergence of deep learning, convolutional neural networks (CNNs) have become the dominant approach for wildfire detection and segmentation. These models can learn complex spatial and spectral patterns directly from data, significantly outperforming traditional methods \cite{hu2021unitemporal, ghali2023deep, rashkovetsky2021wildfire}. Architectures such as U-Net leverage encoder–decoder structures with skip connections to preserve spatial details, while models like DeepLabV3+ incorporate multi-scale contextual information through atrous convolutions and spatial pyramid pooling \cite{ronneberger2015unet, chen2018encoder}.

More recent works have further explored deep learning architectures for active fire detection, including specialized models designed for Landsat imagery \cite{gao2025dafdm}. In parallel, transformer-based architectures have been introduced such as SegFormer and FireFormer highlight the potential of attention-based representations for wildfire segmentation tasks \cite{xie2021segformer, zhang2024fireformer}.

The role of spectral information has also been extensively investigated. Multispectral inputs, particularly those including SWIR bands, have been shown to significantly improve fire detection performance due to their sensitivity to high-temperature emissions and reduced interference from smoke and atmospheric effects \cite{wang2021smoke, ravan2026realtime}. Additionally, selecting appropriate band combinations can be more effective than using all available spectral channels \cite{xu2024sen2fire}.

Another key challenge in wildfire segmentation is the severe imbalance between fire and background pixels, combined with large variability in fire morphology, spatial distribution, and scale \cite{cui2025classbalanced}.

Beyond active-fire detection and segmentation, recent studies have also investigated wildfire spread prediction and calibration using optimization and metaheuristic approaches. Pereira et al. \cite{pereira2022wildfire} explored the calibration of wildfire spread models using evolutionary and stochastic optimization algorithms, while later extending this framework to two-dimensional wildfire propagation modeling under real wildfire scenarios \cite{pereira2024metaheuristic}. These works highlight the inherent variability and uncertainty of wildfire behavior, reinforcing the importance of robust modeling and evaluation strategies.

Despite these advances, the problem of generalization across different fire scales remains largely unexplored. In real-world scenarios, wildfire events evolve dynamically over time, leading to substantial changes in spatial structure and scale \cite{singh2025activewildfire}. However, most existing studies rely on random dataset splits, which do not explicitly evaluate model robustness under distribution shifts.


In this work, we address this gap by introducing a scale-based evaluation protocol, rather than a new segmentation architecture, enabling a controlled analysis of model generalization from small-scale to large-scale fire regions.

\section{Methodology}

This section describes the proposed scale-based wildfire segmentation framework, including the preprocessing pipeline, dataset partitioning strategy, spectral configurations, evaluated segmentation models, and experimental setup.

\begin{figure*}[t]
\centering
    \includegraphics[trim={1cm 0.6cm 1cm 1cm}, width=0.95\textwidth]{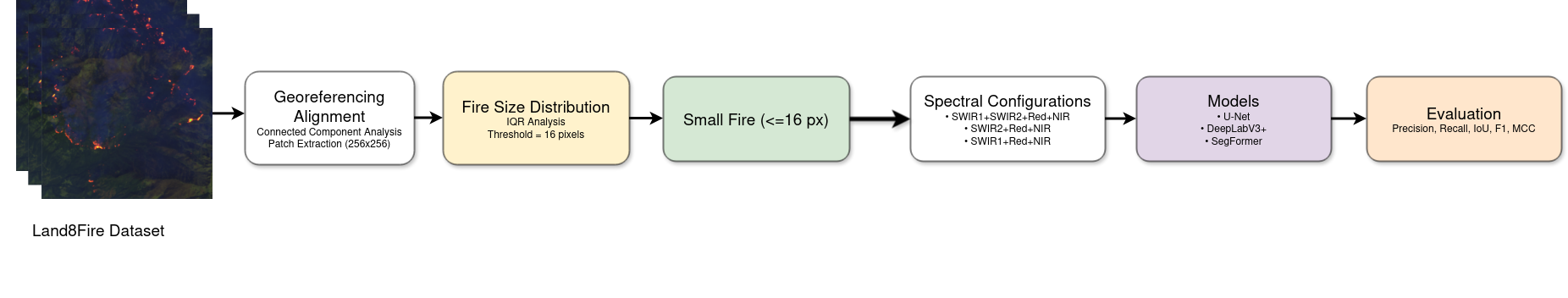}
    \caption{Overview of the proposed scale-based wildfire segmentation pipeline. The raw Land8Fire scenes are first preprocessed through image-mask alignment, patch extraction, and connected-component analysis. Fire-region sizes are then analyzed using the IQR criterion to define small- and large-fire partitions. The resulting data splits are evaluated using different SWIR-based spectral configurations and segmentation models.}
\label{fig:pipeline1}
\end{figure*}

Figure~\ref{fig:pipeline1} summarizes the overall workflow adopted in this study. The pipeline consists of three main stages: preprocessing and fire-region extraction, statistical fire-size analysis, and scale-based dataset partitioning.

\subsection{Dataset}
The dataset used in this work is the Land8Fire \cite{tran2025land8fire}, a large-scale, high-resolution multispectral benchmark designed for semantic segmentation of active wildfires using satellite imagery. It was developed to address limitations of existing wildfire datasets, such as limited spectral diversity, noisy annotations, and severe class imbalance.

Land8Fire is derived from multispectral imagery acquired by the Landsat-8 satellite and extends the previously released ActiveFire dataset \cite{pereira2021activefire} by incorporating manually validated fire masks and improved sampling strategies. The dataset includes wildfire events from multiple regions worldwide, including the Amazon, Africa, Australia, and the United States, providing significant diversity in vegetation types and fire behavior.

Land8Fire provides two versions of the dataset: a preprocessed version and a raw version containing the original selected Landsat-8 scenes. In this work, we use the raw version of the dataset, and all preprocessing steps, including patch extraction and scale-based splitting, are performed using the pipeline described in Section III-B. This allows full control over the data distribution and enables the proposed scale-based analysis.

\subsection{Data Preprocessing and Scale-Based Splitting Pipeline}

Figure~\ref{fig:pipeline1} illustrates the proposed pipeline for evaluating wildfire segmentation under a scale-based distribution shift scenario. The pipeline consists of three main stages: patch extraction and connected-component analysis, statistical fire-size characterization, and scale-based dataset partitioning.

All Landsat-8 multispectral images were first aligned with their corresponding ground-truth masks through georeferencing to ensure spatial consistency between inputs and labels.

Next, each image was divided into non-overlapping patches of size $256 \times 256$ pixels. A connected-component analysis was then applied to the binary fire masks to identify contiguous fire regions within each patch. The size of each connected component was computed as the number of connected fire pixels.


To characterize fire-size variability, the Interquartile Range (IQR) was computed over all detected connected components. The resulting distribution was highly skewed, with most fire regions concentrated in the lower range and a limited number of large outliers. The IQR-based criterion yielded a 16-pixel cutoff, which was used to separate typical small fire components from atypically large connected fire regions. Since each Landsat-8 pixel corresponds to an area of $30 \times 30$ meters, this threshold represents approximately 1.44 hectares.

\begin{figure}[t]
\centering
\includegraphics[width=0.7\linewidth]{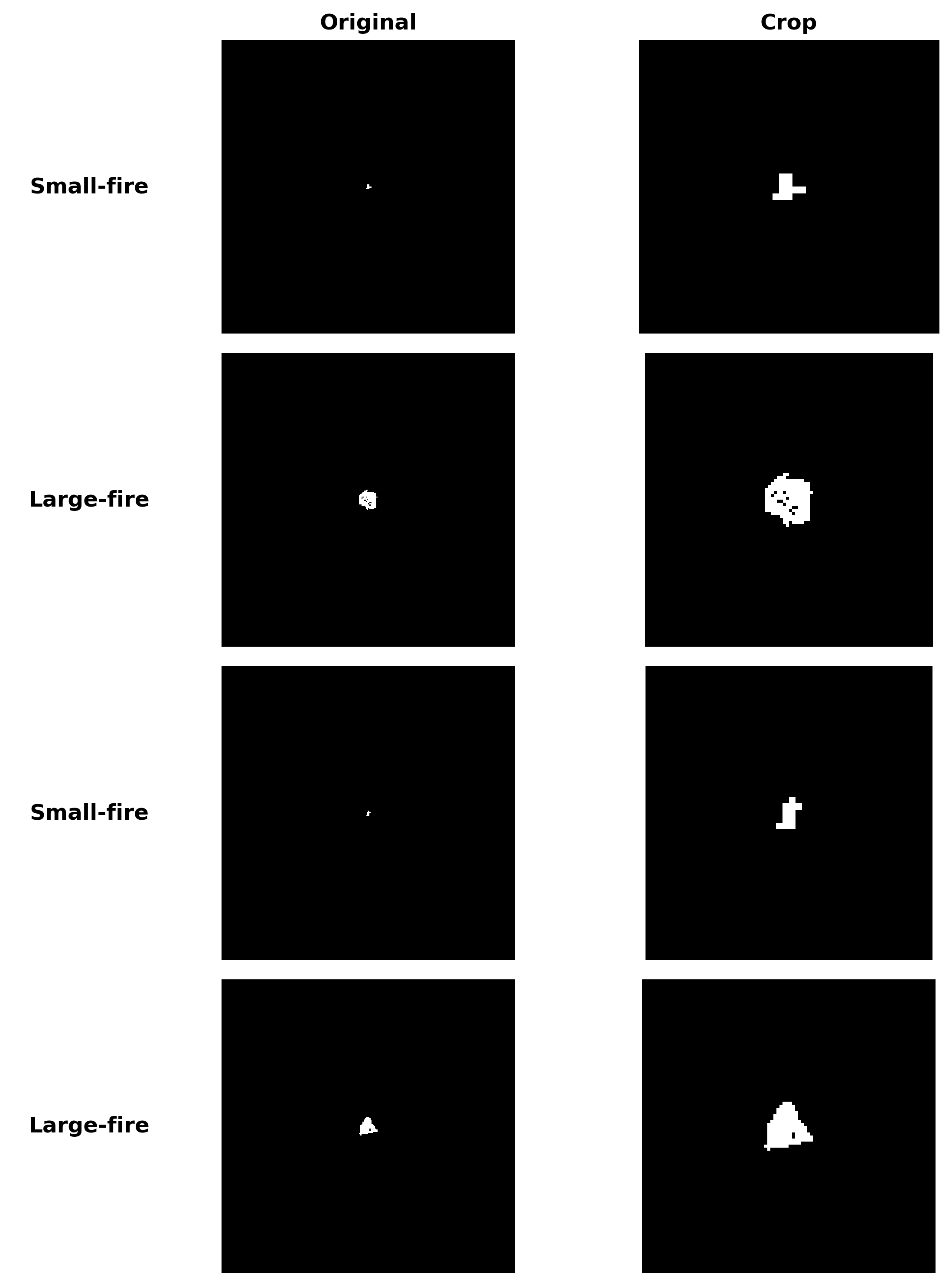}
\caption{Examples of fire masks from the proposed scale-based partitioning. Rows labeled as small-fire contain patches whose connected fire regions remain below the IQR-derived threshold of 16 pixels, whereas rows labeled as large-fire contain patches with at least one connected fire region exceeding this threshold. For each example, the original $256 \times 256$ mask and a cropped view around the fire region are shown.}
\label{fig:small_large_fire_examples}
\end{figure}

In this work, the terms \textit{small-scale} and \textit{large-scale} refer to connected fire-region sizes observed within individual image patches rather than to operational wildfire size classes. Figure~\ref{fig:small_large_fire_examples} presents representative examples of patches categorized as small-fire and large-fire samples according to the proposed criterion.

Dataset partitioning was performed based on the size of connected fire regions identified within each patch:

\begin{itemize}
    \item Patches with largest connected component size $\leq 16$ pixels were assigned to the training set;
    
    \item Patches with largest connected component size $> 16$ pixels were assigned to the test set.
\end{itemize}

This strategy introduces a controlled fire-scale distribution shift, enabling the evaluation of model generalization from small to large fire structures.

Following the Land8Fire curation protocol, patches without fire pixels were excluded from the experiments. The resulting dataset contains approximately 12,000 training patches and 400 testing patches.

\subsection{Spectral Band Configurations}

To evaluate the impact of spectral information on fire detection and generalization under scale-based distribution shifts, three different band combinations were considered. These configurations were designed to analyze how the inclusion or exclusion of shortwave infrared (SWIR) channels affects model performance.

\begin{itemize}
    \item \textbf{Dual-SWIR}: SWIR1, SWIR2, Red, and NIR. This configuration includes both SWIR bands and it is used to evaluate the complementary contribution of SWIR1 and SWIR2.
    \item \textbf{SWIR2-only}: SWIR2, Red, and NIR. This configuration focuses on the longer-wavelength SWIR2 band.
    \item \textbf{SWIR1-only}: SWIR1, Red, and NIR. This configuration focuses on the shorter-wavelength SWIR1 band.
\end{itemize}

SWIR bands are known to be highly responsive to active fire due to their sensitivity to high-temperature emissions and reduced interference from smoke and atmospheric effects. In particular, SWIR2 is more sensitive to higher temperature ranges, while SWIR1 captures complementary thermal and reflectance properties.

By comparing these configurations, we aim to assess:
(i) the individual contribution of each SWIR band,
(ii) whether combining both SWIR bands provides complementary information, and
(iii) how spectral richness influences the ability of models to generalize from small-scale to large-scale fire events.

All configurations maintain the Red and Near-Infrared (NIR) bands, which provide contextual information related to vegetation and background conditions. Thermal bands B10 and B11 were not included because their native spatial resolution is 100 m, requiring resampling to 30 m and potentially introducing spatial uncertainty in patch-level segmentation. This study therefore focuses on 30 m OLI bands to isolate the effect of SWIR information under consistent spatial resolution.

\subsection{Segmentation Models}

Three semantic segmentation architectures were evaluated in this work: U-Net, DeepLabV3+, and SegFormer. These models were selected due to their complementary characteristics regarding spatial-detail preservation, contextual aggregation, and global representation learning.

\subsubsection{U-Net}

U-Net \cite{ronneberger2015unet} is an encoder-decoder convolutional architecture with skip connections that preserve fine-grained spatial information during feature reconstruction. Due to its strong localization capability, U-Net is well suited for detecting sparse and small fire regions.

\subsubsection{DeepLabV3+}

DeepLabV3+ \cite{chen2018encoder} incorporates atrous convolutions and Atrous Spatial Pyramid Pooling (ASPP) to capture multi-scale contextual information. Its large receptive fields enable improved representation of spatially complex fire structures.

\subsubsection{SegFormer}

SegFormer \cite{xie2021segformer} is a transformer-based semantic segmentation architecture that combines hierarchical self-attention encoders with a lightweight multilayer perceptron decoder. Its design enables efficient multi-scale representation learning while preserving spatial structure.

By comparing these three architectures, we analyze how different design trade-offs—local detail preservation (U-Net), contextual multi-scale aggregation (DeepLabV3+), and global attention-based representation learning (SegFormer)—affect generalization from small to large fire events.

\paragraph{Multi-channel Input Adaptation}

Since the original DeepLabV3+ and SegFormer backbones are designed for three-channel (RGB) inputs, they were adapted to support four-channel multispectral data. Specifically, the first convolutional projection layer of each backbone was modified to accept four input channels.

For DeepLabV3+, the weights corresponding to the original three channels were preserved, while the additional channel was initialized using the mean of the pretrained RGB weights. This initialization strategy allows the model to leverage pretrained features while incorporating additional spectral information.

A similar strategy was adopted for SegFormer, extending its input projection layer to handle four-channel multispectral inputs while preserving pretrained initialization as much as possible.

These modifications enable the use of multispectral inputs without significantly altering the original architectures or training dynamics.

\subsection{Evaluation Metrics}

Model performance was evaluated using five standard segmentation metrics: Precision ($P$), Recall ($R$), F1-score ($F_1$), Intersection over Union (IoU), and Matthews Correlation Coefficient (MCC).

\section{Experimental Results}

This section presents the experimental evaluation of the proposed scale-based wildfire segmentation framework. The goal of these experiments is to analyze how different data distributions influence model performance, with a particular focus on generalization across fire scales. We compare a conventional random dataset split, which represents a standard training scenario, with the proposed scale-based splitting strategy that introduces a controlled distribution shift between training and testing data. This comparison allows us to assess whether traditional evaluation protocols provide reliable estimates of model performance in realistic wildfire scenarios. Additionally, we evaluate the behavior of models trained exclusively on small-scale fire regions under different testing conditions, including both similar and significantly different data distributions. This enables a detailed analysis of how scale variability impacts segmentation performance. To evaluate the impact of scale-based distribution shifts on wildfire segmentation, we designed a set of experiments comparing a conventional random split with the proposed scale-based splitting strategy.


\subsection{Training Setup}


All models were trained for binary semantic segmentation, where pixels were classified as either fire or background. To ensure a fair comparison, the same learning rate ($1 \times 10^{-4}$), batch size (8), weight decay ($5 \times 10^{-4}$), and total number of training iterations (5,000) were adopted across all architectures whenever possible. All reported results correspond to the mean and standard deviation over 5 independent runs with different random seeds, using the preprocessing pipeline described in Section III-B. U-Net was trained using weighted cross-entropy loss to mitigate pixel-level class imbalance. DeepLabV3+ employed the deeplabv3plus\_mobilenet backbone with output stride 16 and polynomial learning-rate decay. SegFormer was trained using the same dataset partitions while preserving its default architecture-specific hyperparameters. For DeepLabV3+ and SegFormer, the input projection layers were adapted to support four-channel multispectral inputs by extending the original RGB initialization with an additional channel initialized from the mean pretrained weights. Experiments were conducted on GeForce RTX 3090 24GB.

\subsection{Baseline Setup}

As a reference, a baseline dataset split was created using a standard random partitioning strategy, where the dataset was divided into 70\% for training, 15\% for validation, and 15\% for testing. This split follows common practices in deep learning and does not consider fire size or spatial characteristics. The same preprocessing pipeline described in Section III-B was applied, excluding the scale-based filtering step. This baseline setup represents a conventional training scenario where models are exposed to a mixture of fire sizes during training.

\subsection{Scale-Based Setup}

In contrast, the proposed setup follows the scale-based splitting strategy described in Section III-B. In this case, the training set contains only small-scale fire regions, while the test set consists of patches of large fire regions. This setup introduces a controlled distribution shift, enabling the evaluation of model generalization from small-scale to large-scale fire events.

\subsection{Model Training}

All three models were trained under the two setups described above (baseline and scale-based). The same spectral band configurations and preprocessing pipeline were used across architectures to ensure a consistent comparison.

\subsection{Evaluation Protocol}

To further analyze generalization behavior, the model trained on small-scale fire data (referred to as the \textit{small-fire model}) was evaluated under three different testing conditions:

\begin{itemize}
    \item \textbf{Baseline:} evaluates how the model performs on a standard data distribution containing a mixture of fire sizes.
    \item \textbf{Small-fire:} evaluates performance on data similar to the training distribution.
    \item \textbf{Large-fire:} evaluates the model under a strong distribution shift, where fire regions are significantly larger than those seen during training.
\end{itemize}

This evaluation protocol allows us to analyze how well models generalize across different fire scales and to quantify the impact of scale-based distribution shifts on segmentation performance.

\subsection{Results}


\subsubsection{U-Net Results}

Table~\ref{tab:UnetCombined} reports the performance of U-Net on the baseline, small-fire, and large-fire test sets, respectively. U-Net achieves the strongest overall performance among the evaluated models. On the baseline and large-fire test sets, the Dual-SWIR and SWIR2-only configurations obtain very similar results, with F1-scores above 92\% and IoU values above 86\%. This indicates that U-Net is highly effective at detecting fire regions when trained on small-scale samples and evaluated on larger fire structures. On the small-fire test set, performance decreases, mainly in precision and IoU, while recall remains consistently high across all spectral configurations. This suggests that small and sparse fire regions are more challenging due to their limited spatial extent and higher sensitivity to false positives. Nevertheless, the model maintains high recall, indicating that it rarely misses fire pixels. Across all U-Net experiments, Dual-SWIR and SWIR2-only provide very similar results, while the SWIR1-only configuration performs consistently worse. This suggests that SWIR2 band contains more discriminative information for active fire segmentation than SWIR1 band.


\begin{table}[!b]
\centering
\caption{U-Net performance across baseline, small-fire, and large-fire test sets (mean $\pm$ std)}
\scriptsize
\setlength{\tabcolsep}{1.8pt}
\renewcommand{\arraystretch}{1.12}
\begin{tabular}{cc|ccccc}
\hline
Set & Input & Prec. & Rec. & IoU & F1 & MCC \\
\hline

\multirow{3}{*}{Baseline}
& Dual-SWIR
& 90.26{\tiny$\pm$}.90
& 99.05{\tiny$\pm$}.64
& 89.49{\tiny$\pm$}.81
& 94.45{\tiny$\pm$}.45
& .944{\tiny$\pm$}.004 \\

& SWIR2-only
& 89.04{\tiny$\pm$}.92
& 99.34{\tiny$\pm$}.15
& 88.51{\tiny$\pm$}.80
& 93.91{\tiny$\pm$}.45
& .939{\tiny$\pm$}.004 \\

& SWIR1-only
& 80.72{\tiny$\pm$}1.87
& 95.46{\tiny$\pm$}1.29
& 77.70{\tiny$\pm$}.90
& 87.45{\tiny$\pm$}.57
& .875{\tiny$\pm$}.005 \\
\hline

\multirow{3}{*}{Small-Fire}
& Dual-SWIR
& 88.38{\tiny$\pm$}1.24
& 98.90{\tiny$\pm$}.79
& 87.51{\tiny$\pm$}1.21
& 93.34{\tiny$\pm$}.69
& .934{\tiny$\pm$}.007 \\

& SWIR2-only
& 87.00{\tiny$\pm$}1.23
& 99.23{\tiny$\pm$}.21
& 86.41{\tiny$\pm$}1.08
& 92.71{\tiny$\pm$}.62
& .928{\tiny$\pm$}.006 \\

& SWIR1-only
& 78.11{\tiny$\pm$}1.84
& 98.21{\tiny$\pm$}.56
& 77.01{\tiny$\pm$}1.89
& 87.00{\tiny$\pm$}1.20
& .874{\tiny$\pm$}.011 \\
\hline

\multirow{3}{*}{Large-Fire}
& Dual-SWIR
& 88.52{\tiny$\pm$}.91
& 98.96{\tiny$\pm$}.71
& 87.70{\tiny$\pm$}.98
& 93.44{\tiny$\pm$}.55
& .935{\tiny$\pm$}.005 \\

& SWIR2-only
& 87.13{\tiny$\pm$}1.25
& 99.28{\tiny$\pm$}.17
& 86.58{\tiny$\pm$}1.12
& 92.81{\tiny$\pm$}.64
& .929{\tiny$\pm$}.006 \\

& SWIR1-only
& 78.28{\tiny$\pm$}1.98
& 98.33{\tiny$\pm$}.41
& 77.25{\tiny$\pm$}2.00
& 87.16{\tiny$\pm$}1.26
& .875{\tiny$\pm$}.012 \\
\hline
\end{tabular}
\label{tab:UnetCombined}
\end{table}

\subsubsection{DeepLabV3+ Results}

Table~\ref{tab:DeepLabCombined} presents the results obtained with DeepLabV3+. Compared with U-Net, DeepLabV3+ shows substantially lower performance, especially in recall. Although the model maintains relatively high precision, recall drops significantly across all test sets, indicating that DeepLabV3+ tends to produce conservative predictions. In practice, this means that when DeepLabV3+ predicts fire, it is often correct, but it misses a large portion of actual fire pixels. This behavior is especially evident in the SWIR1-only configuration, where recall and IoU are considerably lower than in the other spectral configurations. The best DeepLabV3+ results are obtained using Dual-SWIR or SWIR2-only, reinforcing the importance of SWIR2 band for active fire detection. Overall, these results suggest that DeepLabV3+ is more sensitive to the scale-based distribution shift than U-Net. Its stronger reliance on contextual feature aggregation may reduce its ability to detect sparse or spatially fragmented fire patterns when trained on small-fire regions.


\begin{table}[!t]
\centering
\caption{DeepLabV3+ performance across baseline, small-fire, and large-fire test sets (mean $\pm$ std)}
\scriptsize
\setlength{\tabcolsep}{1.8pt}
\renewcommand{\arraystretch}{1.12}
\begin{tabular}{cc|ccccc}
\hline
Set & Input & Prec. & Rec. & IoU & F1 & MCC \\ 
\hline

\multirow{3}{*}{Baseline} 
& Dual-SWIR & 75.72{\tiny$\pm$}1.63 & 53.81{\tiny$\pm$}6.56 & 45.89{\tiny$\pm$}5.21 & 62.76{\tiny$\pm$}5.02 & .631{\tiny$\pm$}.045 \\
& SWIR2-only & 76.98{\tiny$\pm$}1.57 & 57.39{\tiny$\pm$}2.85 & 48.99{\tiny$\pm$}2.49 & 65.73{\tiny$\pm$}2.24 & .659{\tiny$\pm$}.021 \\
& SWIR1-only & 79.35{\tiny$\pm$}2.54 & 30.59{\tiny$\pm$}2.57 & 28.32{\tiny$\pm$}2.26 & 44.10{\tiny$\pm$}2.79 & .486{\tiny$\pm$}.023 \\ 
\hline

\multirow{3}{*}{Small-Fire}
& Dual-SWIR & 73.22{\tiny$\pm$}1.68 & 59.16{\tiny$\pm$}5.10 & 48.65{\tiny$\pm$}4.02 & 65.38{\tiny$\pm$}3.75 & .653{\tiny$\pm$}.035 \\
& SWIR2-only & 74.50{\tiny$\pm$}1.87 & 59.85{\tiny$\pm$}3.79 & 49.65{\tiny$\pm$}2.77 & 66.32{\tiny$\pm$}2.42 & .663{\tiny$\pm$}.023 \\
& SWIR1-only & 77.56{\tiny$\pm$}3.78 & 33.86{\tiny$\pm$}3.96 & 30.76{\tiny$\pm$}3.18 & 46.97{\tiny$\pm$}3.78 & .507{\tiny$\pm$}.029 \\
\hline

\multirow{3}{*}{Large-Fire}
& Dual-SWIR & 73.49{\tiny$\pm$}1.32 & 58.84{\tiny$\pm$}5.00 & 48.53{\tiny$\pm$}3.75 & 65.27{\tiny$\pm$}3.50 & .653{\tiny$\pm$}.032 \\
& SWIR2-only & 74.26{\tiny$\pm$}1.44 & 59.90{\tiny$\pm$}3.10 & 49.58{\tiny$\pm$}2.25 & 66.27{\tiny$\pm$}1.98 & .662{\tiny$\pm$}.019 \\
& SWIR1-only & 77.78{\tiny$\pm$}2.58 & 33.58{\tiny$\pm$}2.86 & 30.60{\tiny$\pm$}2.27 & 46.83{\tiny$\pm$}2.70 & .506{\tiny$\pm$}.021 \\

\hline
\end{tabular}
\label{tab:DeepLabCombined}
\end{table}


\subsubsection{SegFormer Results}

Table~\ref{tab:SegFormerCombined} shows the performance of SegFormer. SegFormer achieves intermediate performance between U-Net and DeepLabV3+. On the baseline and large-fire test sets, SegFormer performs strongly with Dual-SWIR and SWIR2-only, achieving F1-scores above 85\% and IoU values above 74\%. However, performance decreases substantially when SWIR1-only is used, particularly in recall and IoU. On the small-fire test set, SegFormer shows a clear reduction in IoU and F1-score compared with the baseline and large-fire sets. This confirms that small fire regions are more difficult to segment, even for transformer-based architectures. However, unlike DeepLabV3+, SegFormer maintains a more balanced precision-recall trade-off, suggesting better robustness to small and sparse fire patterns. The results indicate that SegFormer benefits from the inclusion of both SWIR bands, but SWIR2-only remains highly competitive. This trend is consistent with the U-Net and DeepLabV3+ results.

\begin{table}[h!]
\centering
\caption{SegFormer performance across baseline, small-fire, and large-fire test sets (mean $\pm$ std)}
\scriptsize
\setlength{\tabcolsep}{1.8pt}
\renewcommand{\arraystretch}{1.12}
\begin{tabular}{cc|ccccc}
\hline
Set & Input & Prec. & Rec. & IoU & F1 & MCC \\ 
\hline

\multirow{3}{*}{Baseline} 
& Dual-SWIR & 87.65{\tiny$\pm$}.92 & 87.98{\tiny$\pm$}1.09 & 78.27{\tiny$\pm$}.32 & 87.81{\tiny$\pm$}.20 & .876{\tiny$\pm$}.002 \\
& SWIR2-only & 87.65{\tiny$\pm$}.39 & 87.88{\tiny$\pm$}.37 & 78.19{\tiny$\pm$}.32 & 87.76{\tiny$\pm$}.20 & .875{\tiny$\pm$}.002 \\
& SWIR1-only & 87.04{\tiny$\pm$}.47 & 71.45{\tiny$\pm$}.74 & 64.58{\tiny$\pm$}.60 & 78.48{\tiny$\pm$}.44 & .785{\tiny$\pm$}.004 \\ 
\hline

\multirow{3}{*}{Small-Fire}
& Dual-SWIR & 85.33{\tiny$\pm$}1.12 & 85.53{\tiny$\pm$}.98 & 74.55{\tiny$\pm$}.35 & 85.42{\tiny$\pm$}.23 & .852{\tiny$\pm$}.002 \\
& SWIR2-only & 85.70{\tiny$\pm$}.46 & 85.31{\tiny$\pm$}.50 & 74.68{\tiny$\pm$}.48 & 85.50{\tiny$\pm$}.32 & .853{\tiny$\pm$}.003 \\
& SWIR1-only & 86.23{\tiny$\pm$}.39 & 73.52{\tiny$\pm$}.86 & 65.80{\tiny$\pm$}.77 & 79.37{\tiny$\pm$}.56 & .793{\tiny$\pm$}.005 \\
\hline

\multirow{3}{*}{Large-Fire}
& Dual-SWIR & 85.66{\tiny$\pm$}.95 & 85.46{\tiny$\pm$}1.11 & 74.75{\tiny$\pm$}.31 & 85.55{\tiny$\pm$}.20 & .853{\tiny$\pm$}.002 \\
& SWIR2-only & 85.93{\tiny$\pm$}.48 & 85.27{\tiny$\pm$}.32 & 74.82{\tiny$\pm$}.27 & 85.60{\tiny$\pm$}.18 & .854{\tiny$\pm$}.002 \\
& SWIR1-only & 86.39{\tiny$\pm$}.33 & 73.61{\tiny$\pm$}.68 & 65.96{\tiny$\pm$}.45 & 79.49{\tiny$\pm$}.33 & .795{\tiny$\pm$}.003 \\

\hline
\end{tabular}
\label{tab:SegFormerCombined}
\end{table}

\subsubsection{Cross-Model Analysis}


The results reveal three main findings. First, U-Net provides the most robust performance across all evaluated conditions, especially in terms of recall, IoU, F1-score, and MCC. This behavior can be attributed to its encoder-decoder structure with skip connections, which helps preserve fine spatial details and supports dense pixel-level localization. Such properties are particularly relevant for active-fire masks, where target regions are sparse, fragmented, and highly imbalanced.

Second, DeepLabV3+ exhibits the weakest generalization behavior, mainly due to low recall. Its predictions are more conservative, leading to fewer false positives but more missed fire pixels. This is undesirable for wildfire monitoring, where missed detections may delay response actions. Third, the spectral configuration strongly affects performance across all architectures. SWIR2-only consistently outperforms SWIR1-only, while the Dual-SWIR configuration generally provides the most stable results. These findings support the importance of SWIR information, particularly SWIR2, for active wildfire mapping. Overall, the experiments show that both architecture choice and spectral band selection are critical for robust wildfire segmentation under scale-based distribution shifts.

\section{Conclusion}

This work investigated wildfire segmentation under fire-scale distribution shifts using the Land8Fire dataset. To this end, we proposed a scale-based dataset partitioning strategy derived from the interquartile range of connected fire-region sizes. Experimental results demonstrated that model robustness strongly depends on both architectural design and spectral configuration. Among the evaluated models, U-Net achieved the strongest overall generalization performance, while SegFormer showed competitive robustness under shifted fire-scale distributions. In contrast, DeepLabV3+ exhibited substantially lower recall, indicating reduced robustness under this evaluation setting.

Across all architectures, SWIR2-only consistently provided the strongest or near-best results, reinforcing the importance of SWIR bands information for active wildfire segmentation in Landsat-8 imagery. Overall, the proposed scale-based evaluation protocol provides a more challenging and informative benchmark than conventional random train-test splits, revealing model behaviors that may remain hidden under standard evaluation settings. Future work includes assessing sensitivity to the scale threshold, incorporating statistical significance testing across repeated runs, and validating the proposed protocol on additional datasets and sensors to evaluate its robustness under broader distribution shifts, including biome variability, seasonal changes, and cross-sensor domain adaptation scenarios.


\bibliographystyle{IEEEtran}
\balance
\bibliography{ref.bib}
\balance

\end{document}